\documentclass{article} 
\usepackage{iclr2027_conference,times}

\usepackage{amsmath,amsfonts,bm}

\def\eqref#1{equation~\ref{#1}}

\def\1{\bm{1}}

\def\va{{\bm{a}}}
\def\vb{{\bm{b}}}

\def\vg{{\bm{g}}}

\def\vp{{\bm{p}}}

\def\vr{{\bm{r}}}

\def\vz{{\bm{z}}}

\DeclareMathAlphabet{\mathsfit}{\encodingdefault}{\sfdefault}{m}{sl}
\SetMathAlphabet{\mathsfit}{bold}{\encodingdefault}{\sfdefault}{bx}{n}

\usepackage{tikz}
\usepackage{xcolor}
\usetikzlibrary{arrows.meta,positioning,calc}

\definecolor{repblue}{RGB}{222,235,247}
\definecolor{probewarm}{RGB}{252,229,205}
\definecolor{taskgray}{RGB}{238,238,238}
\definecolor{opgray}{RGB}{247,247,247}
\definecolor{outgray}{RGB}{232,232,232}
\definecolor{linegray}{RGB}{75,75,75}

\usepackage{hyperref}
\usepackage{url}
\usepackage{cleveref}
\usepackage{booktabs}
\usepackage{makecell}
\usepackage{multirow}
\usepackage{enumitem}

\title{Loss-Invariant Projections as Passive Probes of Learned Representations}

\author{\qquad\qquad\qquad\qquad Akshay Chandrasekhar$^{*}$ \quad\&\quad~ Pavlo Melnyk$^{*}$ \\
\qquad\qquad\qquad\qquad Independent researcher \qquad\qquad~~ Linköping University \\ 
\qquad\qquad\qquad\qquad \texttt{akschand24@gmail.com} ~\quad~~ \texttt{pavlo.melnyk@liu.se} \\ 
\qquad\qquad\qquad\qquad \qquad\qquad\quad $^{*}$Both authors contributed equally.}

\iclrfinalcopy

\begin{document}

\maketitle
\lhead{Preprint.}

\begin{abstract}

Learned feature representations in neural networks often contain structure beyond that directly used by the final task output.
We study this structure using \emph{passive probes} that apply fixed, untrained,
property-independent projections to representations as they evolve during training.
We motivate this approach through the task of prediction on $\mathcal{S}^2$ where
equivalent vector and Hermitian parameterizations reveal an additional loss-invariant trace coordinate.
This motivates a general construction in which fixed random projections serve as
observers of learned features.
Because the observer is loss-invariant and independent of the property being studied, changes in accessibility reflect changes in the representation relative to
the fixed observer rather than adaptation of the observer itself.
We show that ensembles of passive probes can directly reflect task-relevant
information such as target alignment.
Under our constructions, the accessibility of eventual difficulty evolves
differently across tasks.
It increases during training in the regression tasks of surface-normal estimation
and image inpainting but remains near its initial level in image classification.
Comparisons with learned linear probes further show that recoverability and
passive accessibility can evolve differently during training.
Together, these results show how passive probes can separately characterize changes in representation geometry and the accessibility of eventual task difficulty.

    \end{abstract}

\section{Introduction}
\label{sec:introduction}
Learned representations are commonly analyzed by asking what properties can be
recovered from them or how their geometry changes across layers, models, or
training. A standard approach to the first question is \emph{probing} where an
auxiliary predictor is trained on a frozen representation to recover a property
of interest \citep{alain2017understanding,conneau-etal-2018-cram}.
Linear evaluation of frozen features is likewise widely used to evaluate
learned visual representations \citep{pmlr-v119-chen20j}. These approaches
provide useful measures of \emph{recoverability}, which indicates whether an observer adapted to the representation can extract a target property from it.

The interpretation of probe performance, however, depends partly on how the
observer is fitted. Probe capacity can allow memorization that obscures
what can be attributed to the representation \citep{hewitt-liang-2019-designing}, while alternative formulations emphasize sample efficiency, description length, and the inductive bias provided by a representation \citep{voita-titov-2020-information,immer-etal-2022-probing}.
More broadly, different probing methodologies address different questions about a representation \citep{belinkov-2022-probing}.
These observations do not make learned probes unsuitable but clarify that
probe performance reflects both the representation and the observer fitted
to the property of interest.

A complementary line of work compares representations through their geometry.
Representational similarity analysis compares patterns of pairwise dissimilarity \citep{kriegeskorte2008}, while methods based on canonical correlation analysis, such as SVCCA~\citep{raghu2017}, compare representations under transformations of their feature coordinates. 
Centered kernel alignment (CKA) provides a related similarity measure with specified invariances and has proved effective for comparing neural representations across networks and initializations \citep{pmlr-v97-kornblith19a}. Such methods characterize important aspects of representation geometry, but do not ask what becomes accessible to an observer that itself remains unchanged during learning.
More generally, different representational similarity measures encode different notions of equivalence and can differ in their sensitivity to changes in the underlying representation \citep{williams2021,ding2021}.
Such methods characterize important aspects of representation geometry, but
do not ask what becomes accessible to an observer that itself remains
unchanged during learning.

This leaves a complementary observational question:
\emph{what becomes accessible to a fixed observer as the representation changes
during learning?} We study this question using \emph{passive probes} which are loss-invariant, scalar projections sampled independently of the property of interest and held fixed across the training trajectory. Unlike a learned probe, a passive probe is not optimized to recover the target. 
Consequently, changes in its relationship to that target reflect changes in the
representation relative to the fixed observer rather than adaptation of the observer. 
We refer to this quantity as \emph{passive accessibility}, distinguishing it from recoverability after fitting a learned probe.

\begin{figure*}[t]
    \centering

    \begin{tikzpicture}[
        >=Latex,
        font=\small,
        arrow/.style={
            ->,
            >=Latex,
            draw=linegray,
            line width=0.65pt
        },
        rep/.style={
            draw=linegray,
            fill=repblue,
            rounded corners=2pt,
            minimum height=7.5mm,
            align=center,
            inner xsep=5pt,
            inner ysep=2pt
        },
        observer/.style={
            draw=linegray,
            fill=probewarm,
            rounded corners=2pt,
            minimum height=7.5mm,
            align=center,
            inner xsep=5pt,
            inner ysep=2pt
        },
        task/.style={
            draw=linegray,
            fill=taskgray,
            rounded corners=2pt,
            minimum height=7.5mm,
            align=center,
            inner xsep=5pt,
            inner ysep=2pt
        },
        operation/.style={
            draw=linegray,
            fill=opgray,
            rounded corners=2pt,
            minimum height=7.5mm,
            align=center,
            inner xsep=5pt,
            inner ysep=2pt
        },
        output/.style={
            draw=linegray,
            fill=outgray,
            rounded corners=2pt,
            minimum height=7.5mm,
            align=center,
            inner xsep=5pt,
            inner ysep=2pt
        },
        smallnote/.style={
            font=\footnotesize,
            align=center,
            text=linegray
        },
        paneltitle/.style={
            font=\bfseries
        }
    ]


        \node[font=\bfseries] at (6.8,8.25)
            {Task model at a checkpoint};

        \node[task, minimum width=10mm] (x) at (4.35,7.25)
            {$x$};

        \node[rep, minimum width=22mm] (h) at (6.80,7.25)
            {$h(x)$};

        \node[task, minimum width=17mm] (yhat) at (9.25,7.25)
            {$Wh(x)$};

        \draw[arrow] (x) -- node[
        above,
        font=\footnotesize
        ] {$f_{\theta}$} (h);

        \draw[arrow] (h) -- node[
        above,
        font=\footnotesize
        ] {$W$} (yhat);


        \begin{scope}[shift={(0,0)}]

            \node[paneltitle] at (2.0,5.20)
                {(a) Learned probe};

            \node[rep, minimum width=29mm] (traindata) at (2.0,4.25)
                {$\{(h(x_i),e_i)\}_{i=1}^{N}$};

            \node[operation, minimum width=15mm] (fit) at (2.0,3.25)
                {fit};

            \node[observer, minimum width=19mm] (bt) at (2.0,2.25)
                {learned $(\vb, c)$};

            \draw[arrow] (traindata) -- (fit);
            \draw[arrow] (fit) -- (bt);

            \node[rep, minimum width=18mm] (ha) at (0.95,1.15)
                {$h(x)$};

            \node[operation, minimum width=15mm] (dota) at (2.0,0.20)
                {$\langle\cdot,\cdot\rangle$};

            \draw[arrow] (ha) -- (dota);
            \draw[arrow] (bt) -- (dota);

            \node[output, minimum width=25mm] (outa) at (2.0,-0.80)
                {$\vb^\top h(x) + c$};

            \draw[arrow] (dota) -- (outa);

            \node[smallnote] at (2.0,-1.60)
                {property-dependent\\observer};

        \end{scope}


        \begin{scope}[shift={(4.45,0)}]

            \node[paneltitle] at (2.0,5.20)
                {(b) Passive probe};

            \node[rep, minimum width=18mm] (hb) at (0.90,2.95)
                {$h(x)$};

            \node[observer, minimum width=18mm] (ab) at (3.10,2.95)
                {fixed $\va_j$};

            \node[operation, minimum width=15mm] (dotb) at (2.0,1.75)
                {$\langle\cdot,\cdot\rangle$};

            \draw[arrow] (hb) -- (dotb);
            \draw[arrow] (ab) -- (dotb);

            \node[output, minimum width=32mm] (outb) at (2.0,0.60)
                {$q_{j}(x)=\va_j^\top h(x)$};

            \draw[arrow] (dotb) -- (outb);

            \node[smallnote] at (2.0,-0.35)
                {fixed, property-independent\\observer};

        \end{scope}


        \begin{scope}[shift={(9.05,0)}]

            \node[paneltitle] at (2.25,5.20)
                {(c) Strict passive probe};

            \node[task, minimum width=26mm] (WT) at (2.25,4.25)
                {final task readout $W$};

            \node[operation, minimum width=43mm] (PT) at (2.25,3.15)
                {$P_W=W^\top(WW^\top)^{+}W$};

            \draw[arrow] (WT) -- (PT);

            \node[rep, minimum width=18mm] (hc) at (0.70,1.90)
                {$h(x)$};

            \node[observer, minimum width=18mm] (ac) at (3.80,1.90)
                {fixed $\va_j$};

            \node[operation, minimum width=29mm] (hproj) at (0.70,0.65)
                {$(I-P_W)h(x)$};

            \node[operation, minimum width=24mm] (aproj) at (3.80,0.65)
                {$(I-P_W)\va_j$};

            \draw[arrow] (hc) -- (hproj);
            \draw[arrow] (ac) -- (aproj);

            \node[
                operation,
                minimum width=11mm,
                minimum height=6.5mm,
                inner xsep=3pt
            ] (hnorm) at (0.70,-0.45)
                {$\displaystyle\frac{\cdot}{\lVert\cdot\rVert}$};

            \node[
                operation,
                minimum width=11mm,
                minimum height=6.5mm,
                inner xsep=3pt
            ] (anorm) at (3.80,-0.45)
                {$\displaystyle\frac{\cdot}{\lVert\cdot\rVert}$};

            \draw[arrow] (hproj) -- (hnorm);
            \draw[arrow] (aproj) -- (anorm);

            \node[rep, minimum width=20mm] (htilde) at (0.70,-1.55)
                {$\tilde h_{\bot}(x)$};

            \node[observer, minimum width=18mm] (atilde) at (3.80,-1.55)
                {$\tilde{\va}_{j,\bot}$};

            \draw[arrow] (hnorm) -- (htilde);
            \draw[arrow] (anorm) -- (atilde);

            \node[operation, minimum width=15mm] (dotc) at (2.25,-2.60)
                {$\langle\cdot,\cdot\rangle$};

            \draw[arrow] (htilde) -- (dotc);
            \draw[arrow] (atilde) -- (dotc);

            \node[output, minimum width=43mm] (outc) at (2.25,-3.65)
                {$q^{\mathrm{strict}}_{j}(x)
            =(\tilde{\va}_{j,\bot})^\top\tilde h_{\bot}(x)$};

            \draw[arrow] (dotc) -- (outc);

        \end{scope}

    \end{tikzpicture}

    \caption{
        \textbf{Learned and passive observers of a changing representation.}
        At a checkpoint, $f_{\theta}$ maps input $x$ to representation
        $h(x)$, and the task readout $W$ produces prediction $Wh(x)$.
        \textbf{(a)} A learned probe $(\vb, c)$ is fitted from
        representation--property pairs
        $\{(h(x_i),e_i)\}_{i=1}^{N}$, where $e_i$ is the property of interest.
        \textbf{(b)} A passive probe instead uses a fixed,
        property-independent direction $\va_j$, sampled once and never optimized,
        with $q_{j}(x)=\va_j^\top h(x)$.
        \textbf{(c)} The strict construction uses the final task readout $W$
        to define the row-space projector
        $P_W=W^\top(WW^\top)^+W$, projects both $h(x)$ and $\va_j$
        with $I-P_W$, normalizes the resulting vectors, and computes their
        inner product.
    }
    \label{fig:passive_probe_overview}
\end{figure*}
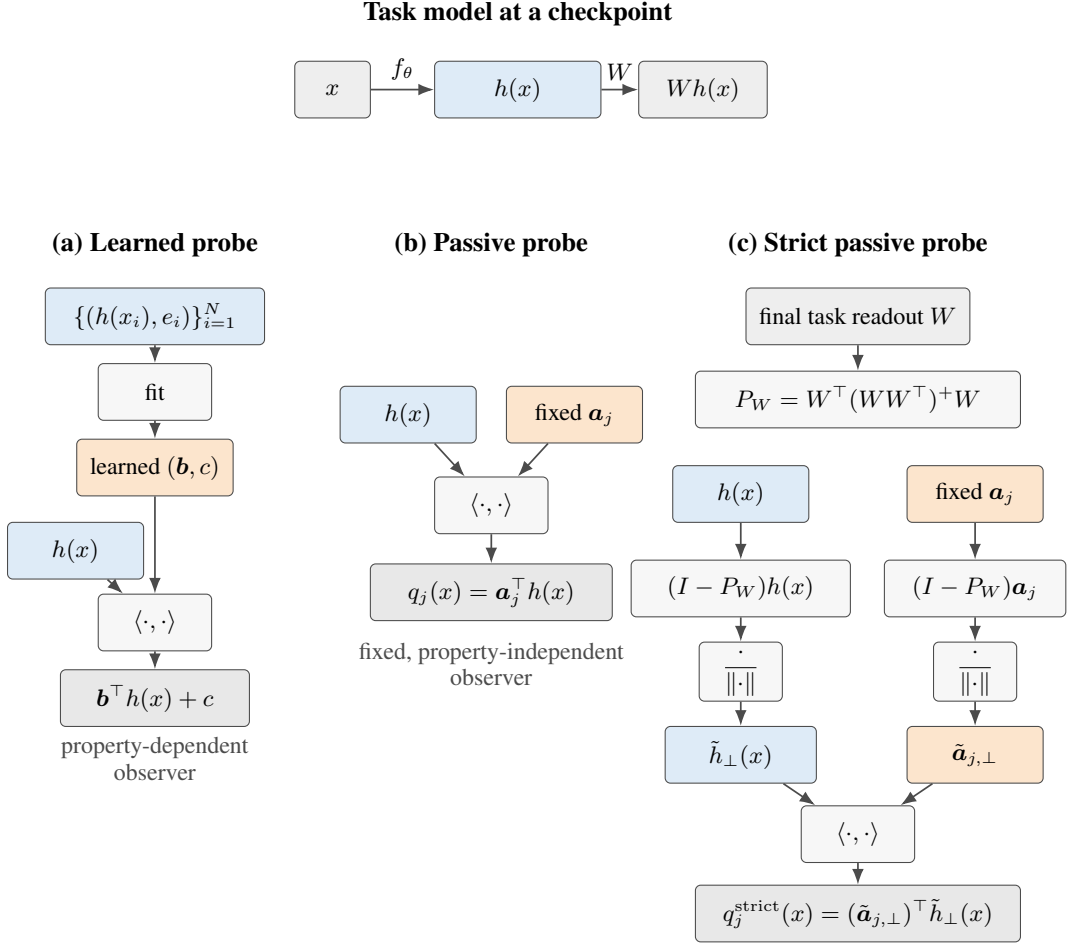

We additionally introduce a strict construction that removes the row space of
the final task readout from the passive direction (see Figure~\ref{fig:passive_probe_overview}). 
Null-space projections have previously been used to remove linearly recoverable properties from representations and to study their behavioral relevance
\citep{ravfogel-etal-2020-null,elazar-etal-2021-amnesic}. 
The role of the null space here is different as we do not learn a direction from the property and then remove it. 
Instead, the passive direction is property-independent and is projected into the null space of the \emph{task readout}. The resulting scalar coordinate is therefore task-readout-invariant. When the loss depends only on that readout, it is also loss-invariant.

Our construction arises naturally from an apparent redundant degree of
freedom in an equivalent parameterization of prediction on the sphere
$\mathcal{S}^2$. This motivating example, presented in Section~\ref{sec:motivating_example}, leads to a general framework for tracking
two aspects of learned representations: changes in their geometry and changes
in the accessibility of a property along fixed scalar projections. 
These quantities need not evolve together. 
Likewise, increasing recoverability by an adapted linear observer need not imply increasing accessibility to an observer that remains fixed.

Our contributions are summarized as follows:

\begin{enumerate}[label=(\roman*)]

\item We introduce \emph{passive probes}: fixed, untrained, and
property-independent scalar observers that track changes in accessibility
without adapting the observer as the representation evolves.

\item We develop a ``strict'' probe construction by restricting passive
directions to the null space of the final task readout, and derive
relationships between passive-probe ensembles and target alignment.

\item Across surface-normal estimation, image classification, and image
inpainting, we find distinct trajectories of passive accessibility to eventual
task difficulty which differ between regression and classification settings.
Comparisons with learned linear probes further show that linear recoverability
and passive accessibility can evolve differently during training.

\end{enumerate}

\section{Methodology}
\label{sec:method}

\subsection{Motivating example: a loss-invariant variable in $\mathcal{S}^2$ prediction}

\label{sec:motivating_example}
Consider the task of predicting a point on the 3D unit sphere $\mathcal{S}^2$. The simplest learning representation is a vector $\vr\in\mathbb{R}^3$ which is subsequently normalized to be on the sphere. We can alternatively represent the point indirectly by the vector $\bm{\nu} \in \mathbb{R}^4$ representing the four unique elements of a $2\times2$ Hermitian matrix
\begin{equation}
    \mathcal{H}
    =
    \begin{pmatrix}
        \nu_1 & \nu_2 + \nu_3 i \\
        \nu_2 - \nu_3 i & \nu_4
    \end{pmatrix}~.
    \label{eq:hermitian}
\end{equation}
From $\mathcal{H}$, the point on $\mathcal{S}^2$ is obtained by extracting the eigenvector of $\mathcal{H}$ corresponding to its smallest eigenvalue. This vector is identified as an element of $\mathbb{CP}^1$ which is isomorphic to $\mathcal{S}^{2}$ (its bijection is given by the stereographic projection). Although the two representations appear quite different, they are known to be algebraically equivalent. More directly,
\begin{equation}
    \vr =
    \left(
        -\nu_2,\,
        \nu_3,\,
        \frac{\nu_4 - \nu_1}{2}
    \right), \qquad
    \|\vr\| = \frac{\lambda_2 - \lambda_1}{2}~. \label{eq:r_equiv}
\end{equation}
where $\lambda_i$ are the eigenvalues of $\mathcal{H}$ with $\lambda_1 \le \lambda_2$. Furthermore, their learning can be shown to be identical under matched parameterizations (see Appendix~\ref{app:proof_equivalence}), with coincident degenerate conditions. The primary difference therefore lies in their dimensionality. The Hermitian representation contains an additional degree of freedom, its trace $\tau = Tr(\mathcal{H}) = \nu_1 + \nu_4$, which is invariant to the loss (see Appendix~\ref{app:proof_trace}). Despite receiving no gradient information, we observe that $\tau$ has the potential to indirectly amass task-relevant information (such as task difficulty) through the learning process. This is because the underlying representation from which it is computed updates during training. This may make the Hermitian representation more favorable over direct prediction of $\vr$. By \cref{eq:r_equiv}, $\vr$ and $\tau$ can be computed as fixed linear projections of the representation $\bm{\nu}$, with the former being task-specific and the latter being loss-invariant. Using this geometric setup as motivation, we analyze the more general case of arbitrary linear projections of learned representations.

\subsection{Linear Projections of Feature Representations}
\label{sec:linear_proj_feat}

Let $h(x)\in\mathbb{R}^{d}$ be a learned feature representation depending on data input $x$, and let $W\in \mathbb{R}^{n \times d}$ be the learned weight matrix corresponding to the final linear layer of a model (see Figure~\ref{fig:passive_probe_overview}, top). Then, model prediction $p(x)$ is simply given by
\begin{equation}
    p(x) = Wh(x).
\end{equation}
We can sample normally distributed linear probes $\va_j$ such that
\begin{equation}
    q_{j}(x) = \va_j^\top h(x).
    \label{eq:passive_probe}
\end{equation}
Note that $\va_j$ is independently constructed, never optimized, and remains fixed throughout training (i.e. $\frac{\partial\mathcal{L}}{\partial \va_j} = 0$). Changes in $q_{j}$ therefore solely arise from changes in the representation $h(x)$ rather than adaptation of the probe. \par

If $q_j(x)$ is constructed from $h(x)$ like $p(x)$, then why would it be able to contain any new task-relevant information not already present in $p(x)$? This is enabled by two main mathematical facts. First, the training loss is applied to $p(x)$ which only ever sees $h_{\parallel}(x)$, the component of $h(x)$ that lies in the row span of $W$. $p(x)$ never sees $h_{\bot}(x)$, the component of $h(x)$ that lies in the nullspace of $W$ (i.e. $Wh_{\bot}(x) = 0$). On the other hand, $q_j$ is derived from both components of $h(x)$, giving it access to information discarded by $W$. This is mathematically described by
\begin{gather}
    p(x) = Wh_{\parallel}(x) + Wh_{\bot}(x) = Wh_{\parallel}(x) ~, \\
    q_j(x) = \va_j^\top h_{\parallel}(x) + \va_j^\top h_{\bot}(x).
\end{gather}
Thus, $q_j$ can be seen as an attempt at recovering more information from the learned representation. Note that $n < d$ for most real-world model architectures, so the dimensionality of $h_{\bot}$ is assumed nonzero. Second, for tasks specifically utilizing an $L_2$ squared loss (e.g. simple regression tasks)\footnote{The same first-moment argument applies to regression constrained to a sphere (i.e. cosine loss). For unit target $\mathcal{Y}$ and unit prediction $\vp$, the conditional expected loss is $1-\vp^\top\mathbb{E}[\mathcal{Y}\mid x]$, so its optimum is the direction of the conditional mean when that mean is nonzero. This does not hold for angular loss.}, we can define the conditional mean and variance $\mu(x)=\mathbb{E}[\mathcal{Y}\mid x]$ and $\Sigma(x)=\operatorname{Cov}(\mathcal{Y}\mid x)$ respectively of the target variable $\mathcal{Y}$, and derive
\begin{gather}
    \mathbb{E}\!\left[\|p(x)-\mathcal{Y}\|^2\mid x\right]
    =
    \|p(x)-\mu(x)\|^2+Tr(\Sigma(x)),
    \\
    \nabla_{p(x)} \mathbb{E}\!\left[\|p(x)-\mathcal{Y}\|^2\mid x\right]
    =
    2(p(x)-\mu(x)).
    \label{eq:squared_loss_conditional}
\end{gather}
The best prediction is therefore $p^*(x)=\mu(x)$. The loss distribution depends on both the conditional mean and variance. However, the gradient only depends on the conditional mean. In other words, the conditional variance, which helps characterize the loss distribution, does not ultimately affect the best prediction. This is in contrast to other loss functions (e.g. cross entropy loss) which may encourage the prediction to incorporate multiple moments of the loss distribution. Both mathematical arguments establish why we would not expect the loss-driven $p(x)$ to contain all the information present in $h(x)$, and the first argument motivates the construction of $q_j$ to potentially capture some of that missing information. However, it is important to clarify that there is no guarantee that $h(x)$ actually contains any additional task-relevant information than what is exposed to $p(x)$ or that this additional information is available via linear probes $\va_j$. Our analyses and findings only lend empirical evidence to the potential accessibilty of additional task information via passive probes. \par

\subsection{Analysis Framework}
\label{sec:analysis_framework}
The theoretical effectiveness of the probes depends entirely on the amount of information available in the feature maps. To further our analysis of this information, we formally define the following quantities
\begin{gather}
    P_W =
    W^\top
    \left(W W^\top\right)^{+}
    W ~,
    \label{eq:task_projector} \\
    h_{\bot}(x) = (I-P_W)h(x)
    \qquad
    \va_{j,\bot} = (I-P_W)\va_j,
    \label{eq:strict_projection}
\end{gather}
where $(\cdot)^{+}$ denotes the Moore--Penrose pseudoinverse. $P_W$ is the projector onto the row space of $W$ and both $h_{\bot}$ and $\va_{j,\bot}$ lie in the nullspace of $W$. These quantities isolate the information present in the probe that is linearly distinct from the information in $p(x)$ and is a useful alternative quantity for comparison in our analysis (see Figure~\ref{fig:passive_probe_overview} and \cref{app:probe_controls}). \par
Next, we attempt to quantify the accessibility of information via the probes (i.e. their potential utility) using metrics from a trained model and a reference dataset. This requires defining the task-relevant information we hope to access. For this paper, we focus on observing task difficulty as defined by the relevant loss function $\mathcal{L}$ (see Section~\ref{sec:experimental_setup}). Note that the following are only diagnostic statistics and not predictive metrics. They can be calculated with either the regular or strict quantities. \par
Given a fixed dataset of $N$ samples $\mathbf{x} = \{x_1, \dots, x_N\}$, we can compute the error vector
\begin{gather}
    \mathbf{e} = \{\mathcal{L}(p(x_1), y_1), \dots, \mathcal{L}(p(x_N), y_N)\}
\end{gather}
from each prediction and target $y_i$ and use it to quantify task difficulty. Applying the matrix operator $C_N=I_N-\mathbf 1_N\mathbf 1_N^\top/N$, we can center error vector as $\bar{\mathbf e}=C_N\mathbf e$.

\paragraph{Linear kernel alignment}
First, the learned features of the dataset are stacked into a matrix
\begin{equation}
    H =
    \begin{bmatrix}
        h(x_1)^\top\\
        \vdots\\
        h(x_N)^\top
    \end{bmatrix}
    \in\mathbb{R}^{N\times d}.
\end{equation}
The Gram entry $K_{i\ell}$ measures the similarity of the selected
features for examples $x_i$ and $x_\ell$ and is computed for the dataset as
\begin{equation}
    K
    =
    HH^\top \qquad
    K_c = C_{N}KC_N.
    \label{eq:kernel_identity}
\end{equation}
Using this similarity, we can define the (normalized) linear alignment metric
\begin{equation}
    A_K
    =
    \frac{
        \bar{\mathbf e}^\top K_c\bar{\mathbf e}
    }{
        \|\bar{\mathbf e}\|^2\operatorname{tr}(K)
    }.
    \label{eq:AK}
\end{equation}
This is the fraction of centered probe-ensemble energy aligned with
eventual difficulty. It is defined when both denominator factors are
nonzero and lies in the range $[0,1]$. Note that $A_K$ is computed independent of the probes $\va_j$ and is thus a property of the learned features and error vector. The similarity kernel is linear, so finite random projections can approximate $A_K$ and access the information it represents.

\paragraph{Nonlinear kernel alignment}
To capture nonlinear relationships with task difficulty, we utilize Gaussian similarity as a radial kernel. Let $s_j>0$ be the standard deviation of $\mathbf{q}_j = H\va_j$, and form
\begin{gather}
    G_{i\ell}
    =
    \frac{1}{m}\sum_{j=1}^m
    \exp\!\left(
              -\frac{(q_j(x_i)-q_j(x_\ell))^2}{2s_j^2}
    \right),
    \qquad
    G_c=C_NGC_N~,
    \\
    A_G
    =
    \frac{
        \bar{\mathbf e}^\top G_c\bar{\mathbf e}
    }{
        \|\bar{\mathbf e}\|^2\operatorname{tr}(G_c)
    }.
    \label{eq:AG}
\end{gather}
$A_G$ measures whether examples that are similar under the random projection kernels have systematically similar centered difficulty. It is similarly only defined for nonzero denominator and lies in range $[0, 1]$. At the population level, it detects variation in mean difficulty along $q_j$,
including non-monotonic variation.
Unlike $A_K$, its value depends on the actual probed values and the number of probes because $G$ is an
average over $m$ scalar Gaussian kernels in probed coordinate space. Increasing $m$ reduces the
Monte Carlo variation of this ensemble statistic.

\paragraph{Statistical measures of correlation}
The previous alignment metrics characterize the accessibility of task difficulty to random projections. To better contextualize them, we also compute some standard measures of correlation to analyze the general relationship between learned features and task difficulty. With $\bar{\mathbf q}_j=C_N\mathbf q_j$, we calculate the squared Pearson
correlation for linear association as
\begin{equation}
    \rho_{Pearson,j}^2
    =
    \frac{(\bar{\mathbf q}_j^\top\bar{\mathbf e})^2}
    {\|\bar{\mathbf q}_j\|^2\|\bar{\mathbf e}\|^2},
    \label{eq:pearson}
\end{equation}
For monotonic association, we calculate the absolute value of the Spearman correlation as
\begin{equation}
    |\rho_{\mathrm{Spearman},j}|
    =
    \Biggl |\frac{\sum_{i=1}^N (r_{j,i}-\bar r_j)(s_i-\bar s)}
    {\sqrt{\sum_{i=1}^N(r_{j,i}-\bar r_j)^2}
    \sqrt{\sum_{i=1}^N(s_i-\bar s)^2}}\Biggr |, \label{eq:spearman}
\end{equation}
where $r_{j,i}, s_{i}$ are the indices of the sorted values of $\mathbf{q}_j$ and $\mathbf{e}$ respectively and $\bar{r}_j, \bar{s}$ are the mean values.
Finally, for a one-dimensional predictor $\hat e_{j,i}=g_j(q_j(x_i))$ fitted only on the calibration split, we compute the coefficient of determination
\begin{equation}
    R_j^2
    =
    1-
    \frac{\sum_{i\in\mathrm{test}}(e_i-\hat e_{j,i})^2}
    {\sum_{i\in\mathrm{test}}(e_i-\bar e_{\mathrm{test}})^2}.
    \label{eq:calibrated_r2}
\end{equation}
This tests whether a single fixed coordinate supports reproducible
nonlinear prediction of difficulty on unseen observations.

\section{Experiments and Results}
\label{sec:experiments}

We study passive accessibility during training in three settings:
surface-normal estimation on NYUv2, image classification on CIFAR-100, and
image inpainting on CIFAR-100. For each task, we train five independently
seeded models and evaluate 256 fixed Gaussian passive probes at a predetermined
sequence of checkpoints.

At every checkpoint, we compare the passive coordinates $q_j(x)$ with the corresponding task-difficulty variable defined by the final trained model. Our primary
analysis reports both raw probes and the task-invariant strict probes defined in Appendix~\ref{app:probe_controls}. We report linear alignment
$A_K$ (\cref{eq:AK}), nonlinear alignment $A_G$ (\cref{eq:AG}), squared Pearson correlation (\cref{eq:pearson}), absolute
Spearman correlation (\cref{eq:spearman}), and held-out calibrated $R^2$ (\cref{eq:calibrated_r2}). The first four quantities
are evaluated directly on the test set. For calibrated $R^2$, the scalar
calibration is fitted on the calibration split and evaluated on the held-out
test split. Task performance is reported on the same test data.
Appendix~\ref{app:evaluation_details} specifies probe sampling, the fixed
final readout subspace, and the calibration and test evaluation procedures.

\subsection{Tasks and experimental setup}
\label{sec:experimental_setup}

\paragraph{Surface-normal estimation}

We train the baseline-normal architecture of DSINE
\citep{bae2024dsine} on the NYUv2 dataset \citep{silberman2012indoor}. At each valid
pixel $x$, the model predicts a unit surface normal $\hat n_\theta(x)$ and is
trained with the cosine loss
\begin{equation}
    \ell_{\mathrm{normal}}(x)
    =
    1-\hat n_\theta(x)^\top n^*(x),
    \label{eq:dsine_objective}
\end{equation}
where $n^*(x)$ is the ground-truth unit normal. We probe the 128-dimensional
representation immediately preceding the final convolutional output.
Task difficulty is the angular error of the final model,
$e(x)=\arccos\!\left(\hat n_{\theta}(x)^\top n^*(x)\right)$.

Models are trained for 20 epochs.
The evaluation uses separate
calibration and test sets containing 327 images each, with 83,712 sampled valid
pixels in each split. Full model, training, and data details are provided in
Appendix~\ref{app:surface_normal_details}.

\paragraph{CIFAR-100 classification}

We train a vision transformer~\citep{dosovitskiy2021an} from scratch on CIFAR-100~\citep{alex2009learning}. For image $x$ with label $y$, the training
objective is cross entropy,
\begin{equation}
    \ell_{\mathrm{cls}}(x,y)
    =
    -\log \pi_\theta(y\mid x),
    \label{eq:classification_objective}
\end{equation}
where $\pi_\theta(y\mid x)$ is the probability assigned to the correct class.
We probe the 192-dimensional CLS representation immediately preceding the
100-class linear layer. Eventual difficulty $e(x,y)$ is the
negative log likelihood in \cref{eq:classification_objective} evaluated
under the final model $\theta$.

We train on 45,000 images, reserve 5,000 images for calibration, and use the
official 10,000-image test set only for evaluation. Models are trained for
300 epochs. Architecture, optimization, augmentation, and checkpoint details
are provided in Appendix~\ref{app:classification_details}.

\paragraph{CIFAR-100 inpainting}

We adopt the vision transformer used for the classification task for image inpainting.
The central $16\times16$ region of each $32\times32$ image is masked, corresponding to 16 spatial patches.
A shared linear reconstruction head predicts the RGB
values of each masked $4\times4$ patch from its spatial-token representation.
We probe the 192-dimensional representation immediately preceding this head.

For image $x$ and masked patch $p$, the reconstruction loss is
\begin{equation}
    \ell_{\mathrm{inp}}(x,p)
    =
    \frac{1}{48}
    \left\|
        \hat y_\theta(x,p)-y(x,p)
    \right\|_2^2,
    \label{eq:inpainting_objective}
\end{equation}
where $y(x,p)\in[0,1]^{48}$ contains the RGB values of the target patch.
Training minimizes this error over the 16 masked patches. Eventual difficulty
$e(x,p)$ is the per-patch reconstruction error in \cref{eq:inpainting_objective} evaluated under the final model $\theta$.

We use 45,000 images for training, 5,000 for calibration, and the official
10,000-image test set for evaluation. Each calibration and test image
contributes all 16 masked patches, giving 80,000 calibration observations and
160,000 test observations. Models are trained for 100 epochs. Architecture,
masking, optimization, and checkpoint details are provided in
Appendix~\ref{app:inpainting_details}.

\subsection{Passive accessibility during training}
\label{sec:main_results}

\begin{figure*}[t]
\centering
\includegraphics[width=\textwidth]
{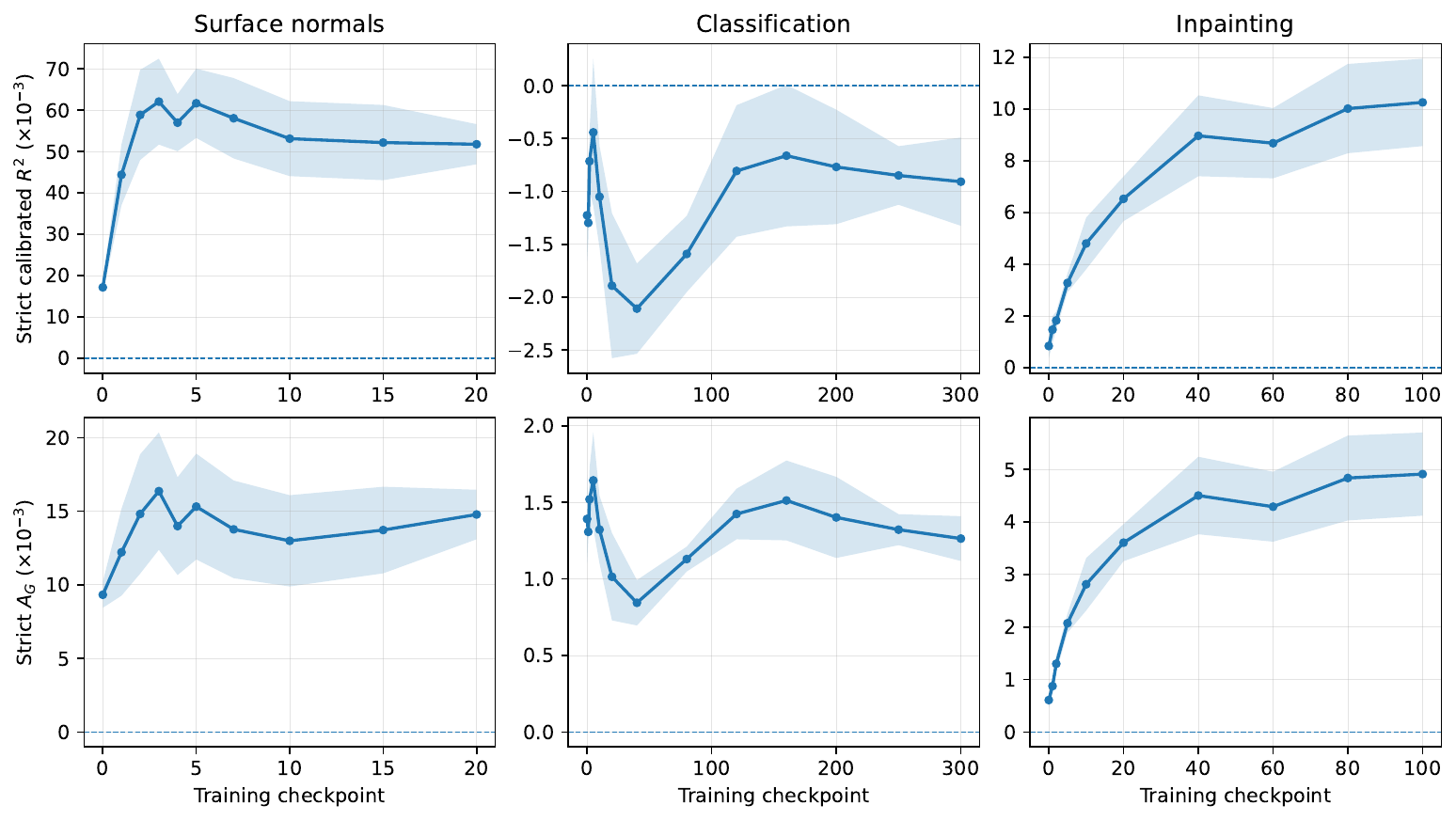}
\caption{
\textbf{Strict passive accessibility across training.}
Calibrated $R^2$ (top) and $A_G$ (bottom) for strict passive probes at each
evaluated training checkpoint, averaged over five independent training seeds;
shading denotes $\pm 1$ standard deviation across seeds.
Surface-normal estimation shows a strong early increase in both measures
followed by an elevated but non-monotonic trajectory, while inpainting shows
a sustained increase across training. Classification exhibits intermediate
variation but no sustained initial-to-final increase in either measure.
Despite their different constructions, the two measures therefore show
consistent task-level patterns. In particular, $A_G$ requires no fitted
mapping from the passive coordinates to eventual difficulty, whereas
calibrated $R^2$ uses calibration-set bin means.
}
\label{fig:strict_accessibility_trajectories}
\end{figure*}

Table~\ref{tab:main_results} summarizes passive accessibility at initialization
and at the final training checkpoint, and corresponding trajectories across all evaluated
checkpoints are reported in
Figure~\ref{fig:strict_accessibility_trajectories}.

\begin{table*}[t]
    \centering
    \caption{
        \textbf{Passive accessibility during training.}
        Entries report initialization $\rightarrow$ final checkpoint, with
        $(\Delta \pm \mathrm{SD}_{\Delta})$ shown below. Initial and final
        values are across-seed means; $\Delta$ is computed within each seed
        before aggregation over five independently trained models. Probe
        metrics are reported in units of $10^{-3}$. Calibrated $R^2$ uses a
        scalar calibration fitted on CAL and evaluated on TEST. Task
        performance is evaluated on TEST and is shared by the raw and strict
        probe families.
    }
    \label{tab:main_results}

    \tiny
    \setlength{\tabcolsep}{2.5pt}
    \renewcommand{\arraystretch}{1.08}

    \begin{tabular}{llccccc@{\hspace{5pt}}c}
        \toprule
        \textbf{Task}
        & \textbf{Family}
        & \multicolumn{5}{c}{\textbf{Probe metrics} ($\times 10^{-3}$)}
        & \textbf{Performance} \\
        \cmidrule(lr){3-7}
        & &
        $\boldsymbol{A_K}$
        & $\boldsymbol{A_G}$
        & \textbf{Pearson$\boldsymbol{^2}$}
        & $\boldsymbol{|\mathrm{Spearman}|}$
        & \textbf{cal.\ $\boldsymbol{R^2}$}
        & \\
        \midrule

        \multirow{2}{*}{Surface normals}
        & Raw
        & \makecell{23.60 $\rightarrow$ 30.87\\$(+7.27 \pm 5.28)$}
        & \makecell{14.55 $\rightarrow$ 21.57\\$(+7.01 \pm 1.19)$}
        & \makecell{18.24 $\rightarrow$ 25.81\\$(+7.56 \pm 3.10)$}
        & \makecell{131.26 $\rightarrow$ 191.87\\$(+60.61 \pm 11.05)$}
        & \makecell{27.91 $\rightarrow$ 64.28\\$(+36.37 \pm 3.98)$}
        & \multirow{2}{*}{
            \makecell{
                75.58 $\rightarrow$ 19.20\\
                $(-56.38 \pm 21.70)$\\
                deg
            }
        }
        \\

        \addlinespace[2pt]

        & Strict
        & \makecell{16.79 $\rightarrow$ 16.00\\$(-0.79 \pm 2.04)$}
        & \makecell{9.33 $\rightarrow$ 14.78\\$(+5.45 \pm 1.33)$}
        & \makecell{13.79 $\rightarrow$ 15.60\\$(+1.81 \pm 1.50)$}
        & \makecell{110.84 $\rightarrow$ 154.95\\$(+44.11 \pm 8.80)$}
        & \makecell{17.13 $\rightarrow$ 51.76\\$(+34.63 \pm 4.61)$}
        & \\

        \addlinespace[5pt]

        \multirow{2}{*}{Classification}
        & Raw
        & \makecell{2.47 $\rightarrow$ 2.37\\$(-0.11 \pm 0.37)$}
        & \makecell{1.41 $\rightarrow$ 1.39\\$(-0.03 \pm 0.22)$}
        & \makecell{1.88 $\rightarrow$ 2.31\\$(+0.43 \pm 0.38)$}
        & \makecell{69.70 $\rightarrow$ 46.78\\$(-22.92 \pm 3.97)$}
        & \makecell{$-1.07 \rightarrow -0.23$\\$(+0.84 \pm 0.47)$}
        & \multirow{2}{*}{
            \makecell{
                1.00 $\rightarrow$ 70.62\\
                $(+69.62 \pm 0.56)$\\
                \%
            }
        }
        \\

        \addlinespace[2pt]

        & Strict
        & \makecell{2.28 $\rightarrow$ 2.02\\$(-0.25 \pm 0.53)$}
        & \makecell{1.39 $\rightarrow$ 1.26\\$(-0.13 \pm 0.37)$}
        & \makecell{1.87 $\rightarrow$ 2.12\\$(+0.25 \pm 0.59)$}
        & \makecell{69.27 $\rightarrow$ 44.20\\$(-25.06 \pm 7.40)$}
        & \makecell{$-1.23 \rightarrow -0.91$\\$(+0.32 \pm 0.62)$}
        & \\

        \addlinespace[5pt]

        \multirow{2}{*}{Inpainting}
        & Raw
        & \makecell{0.62 $\rightarrow$ 4.50\\$(+3.88 \pm 0.87)$}
        & \makecell{0.61 $\rightarrow$ 2.97\\$(+2.36 \pm 0.55)$}
        & \makecell{0.68 $\rightarrow$ 4.84\\$(+4.16 \pm 0.92)$}
        & \makecell{26.94 $\rightarrow$ 76.69\\$(+49.75 \pm 8.05)$}
        & \makecell{0.81 $\rightarrow$ 5.34\\$(+4.54 \pm 1.11)$}
        & \multirow{2}{*}{
            \makecell{
                5.71 $\rightarrow$ 17.31\\
                $(+11.60 \pm 0.01)$\\
                dB
            }
        }
        \\

        \addlinespace[2pt]

        & Strict
        & \makecell{0.63 $\rightarrow$ 9.35\\$(+8.73 \pm 1.61)$}
        & \makecell{0.61 $\rightarrow$ 4.91\\$(+4.30 \pm 0.79)$}
        & \makecell{0.70 $\rightarrow$ 8.60\\$(+7.90 \pm 1.29)$}
        & \makecell{27.79 $\rightarrow$ 106.17\\$(+78.38 \pm 7.75)$}
        & \makecell{0.84 $\rightarrow$ 10.26\\$(+9.42 \pm 1.61)$}
        & \\

        \bottomrule
    \end{tabular}
\end{table*}

The three tasks exhibit distinct patterns. For surface-normal estimation,
passive accessibility increases under several measures, including under the
strict probe construction. In particular, strict $A_G$ and
held-out calibrated $R^2$ increase substantially, whereas strict $A_K$ changes
little. Thus, the nonlinear and calibrated measures reveal a systematic change
that is weak under linear alignment.

For CIFAR-100 classification, task performance improves substantially while
most passive accessibility measures remain near their initialization levels
under both raw and strict probes. Absolute Spearman correlation instead
decreases. Improved task performance therefore does not by itself imply
increased accessibility of eventual difficulty to a fixed,
property-independent scalar observer.

For CIFAR-100 inpainting, passive accessibility increases under both raw and
strict probes, with positive endpoint changes in $A_K$, $A_G$, Pearson$^2$,
absolute Spearman correlation, and held-out calibrated $R^2$. Together, the
three tasks show that passive accessibility can develop substantially during
training, but need not do so simply because task performance improves. It is interesting to note that in our experiments, only the regression loss tasks (which Section~\ref{sec:method} mentioned does not require incorporation of further information on the loss distribution) provided a loss landscape such that task difficulty was accessible through the learned representations.

These task-level patterns also hold when feature normalization and removal
of the final task-readout row space are applied separately, as shown by the
unit-normalized and readout-orthogonal controls in
Appendix~\ref{app:probe_controls}. The calibrated-$R^2$ endpoint conclusions
are also stable across $B\in\{5,10,20,40\}$ calibration quantile bins
(Appendix~\ref{app:bin_ablation}), including the primary choice $B=20$.
\subsection{Passive accessibility and learned recoverability}
\label{sec:learned_probe_comparison}

Passive and learned probes ask different questions. A passive probe holds the
observer fixed throughout training, whereas a learned probe measures how well
eventual difficulty can be recovered after adapting an observer to the
representation at each checkpoint. The two measurements therefore need not
evolve together.

For the learned comparison, we fit a ridge regression from the raw
representation to the same eventual-difficulty target used in the passive
analysis. Regularization is selected using calibration data only, and the
fitted probe is evaluated on held-out test observations. Both analyses use the
raw representation and the same CAL/TEST observations. Strict passive probes
serve as a separate task-readout-invariant control and are not included in
this representation-matched comparison. Appendix~\ref{app:learned_probe_details}
provides the feature-standardization, cross-validation, regularization
selection, and matched evaluation protocol.

Table~\ref{tab:learned_probe_comparison} compares the two observers using
squared Pearson correlation, absolute Spearman correlation, and calibrated
$R^2$. Learned-probe correlations are computed directly from its scalar TEST
predictions. For both observers, calibrated $R^2$ uses the same 20-bin scalar
calibration fitted on CAL and evaluated on TEST. As in
Table~\ref{tab:main_results}, values are reported in units of $10^{-3}$.

\begin{table*}[t]
    \centering
    \caption{
        \textbf{Passive accessibility versus learned recoverability.}
        Passive probes are fixed, property-independent scalar observers,
        whereas learned ridge probes are fitted to eventual difficulty
        separately at each checkpoint. Both use the raw representation and
        the same CAL/TEST observations. Entries report initialization
        $\rightarrow$ final checkpoint, with
        $(\Delta \pm \mathrm{SD}_{\Delta})$ shown below, across five training
        seeds. All metrics are reported in units of $10^{-3}$.
    }
    \label{tab:learned_probe_comparison}

    \small
    \setlength{\tabcolsep}{7pt}
    \renewcommand{\arraystretch}{1.08}

    \begin{tabular}{llccc}
        \toprule
        \textbf{Task}
        & \textbf{Observer}
        & \multicolumn{3}{c}{\textbf{Metrics} ($\times 10^{-3}$)} \\
        \cmidrule(lr){3-5}
        & &
        \textbf{Pearson$\boldsymbol{^2}$}
        & $\boldsymbol{|\mathrm{Spearman}|}$
        & \textbf{cal.\ $\boldsymbol{R^2}$} \\
        \midrule

        \multirow{2}{*}{Surface normals}
        & Passive
        & \makecell{18.24 $\rightarrow$ 25.81\\$(+7.56 \pm 3.10)$}
        & \makecell{131.26 $\rightarrow$ 191.87\\$(+60.61 \pm 11.05)$}
        & \makecell{27.91 $\rightarrow$ 64.28\\$(+36.37 \pm 3.98)$}
        \\

        \addlinespace[2pt]

        & Learned
        & \makecell{93.01 $\rightarrow$ 186.94\\$(+93.94 \pm 13.28)$}
        & \makecell{329.68 $\rightarrow$ 517.11\\$(+187.43 \pm 7.68)$}
        & \makecell{97.17 $\rightarrow$ 189.82\\$(+92.65 \pm 9.25)$}
        \\

        \addlinespace[5pt]

        \multirow{2}{*}{Classification}
        & Passive
        & \makecell{1.88 $\rightarrow$ 2.31\\$(+0.43 \pm 0.38)$}
        & \makecell{69.70 $\rightarrow$ 46.78\\$(-22.92 \pm 3.97)$}
        & \makecell{$-1.07 \rightarrow -0.23$\\$(+0.84 \pm 0.47)$}
        \\

        \addlinespace[2pt]

        & Learned
        & \makecell{10.86 $\rightarrow$ 188.56\\$(+177.70 \pm 6.55)$}
        & \makecell{173.96 $\rightarrow$ 511.53\\$(+337.57 \pm 14.12)$}
        & \makecell{5.29 $\rightarrow$ 188.76\\$(+183.48 \pm 9.82)$}
        \\

        \addlinespace[5pt]

        \multirow{2}{*}{Inpainting}
        & Passive
        & \makecell{0.68 $\rightarrow$ 4.84\\$(+4.16 \pm 0.92)$}
        & \makecell{26.94 $\rightarrow$ 76.69\\$(+49.75 \pm 8.05)$}
        & \makecell{0.81 $\rightarrow$ 5.34\\$(+4.54 \pm 1.11)$}
        \\

        \addlinespace[2pt]

        & Learned
        & \makecell{61.56 $\rightarrow$ 200.91\\$(+139.34 \pm 5.01)$}
        & \makecell{310.01 $\rightarrow$ 615.08\\$(+305.07 \pm 11.01)$}
        & \makecell{64.33 $\rightarrow$ 200.65\\$(+136.31 \pm 4.60)$}
        \\

        \bottomrule
    \end{tabular}
\end{table*}

The clearest separation occurs in classification. Learned recoverability
increases strongly across all three measures, whereas passive accessibility
changes little overall: passive Pearson$^2$ and calibrated $R^2$ remain near
their initialization levels, while absolute Spearman correlation decreases.
Thus, substantial growth in linear recoverability need not be accompanied by
greater accessibility to a fixed, property-independent scalar observer.

Surface-normal estimation and inpainting show a complementary pattern. In
both tasks, passive accessibility and learned recoverability increase during
training, although the learned changes are larger across the matched measures.
Together, the three tasks show that linear recoverability and passive
accessibility can evolve differently during training. They therefore
characterize empirically separable properties of the learned representation.
\section{Conclusion}
\label{sec:conclusion}
Passive probes provide a framework for examining how task-relevant information becomes accessible during learning without explicit fitting of the probe to the property of interest. 
Their relationship to representation geometry supports the interpretation of these measurements, including probing the null space of the final task output. Across the three tasks considered, the findings reveal different levels of accessibility of eventual difficulty from its fixed, untrained directions. 
The contrasting behavior of classification and the two regression tasks indicates that this relationship depends on the learning setting. 
Passive probes differ in information content from learned probes and can thus provide complementary evidence about the organization of learned representations.
Further work is needed to establish how the task, training objective, and architecture contribute to the observed differences.

\section*{AI use statement}

In this work, we used generative AI tools to assist in critically evaluating and refining an author-developed conceptual framework and mathematical derivations, checking mathematical claims, refining hypotheses, designing aspects of the experimental and analysis methodology, implementing and debugging analysis code, and interpreting experimental results.
The core research idea, including the initial identification of the loss-invariant degree of freedom, the relevant equivalence, and the foundational mathematical formulation motivating this work, was developed by the authors.
Generative AI was subsequently used to stress-test these ideas, explore their implications and generalizations, identify potential confounds and controls, and assist in developing and validating the resulting experimental analysis.

We did not use generative AI tools to generate or modify the experimental datasets or to fabricate, alter, or selectively choose experimental measurements. Tasks involving translation, qualitative or thematic data analysis, and human-subject research were not applicable to this work.

Additionally, we used generative AI tools for brainstorming, literature search and synthesis, critical assessment of novelty and positioning, software development and debugging, organization of the manuscript, and drafting and editing scientific text for clarity.

All AI-assisted research contributions were reviewed and validated by the authors. Mathematical statements and derivations used in the work were checked by the authors against their definitions and where applicable verified numerically.
AI-assisted experimental and analysis code was tested using analytic or synthetic sanity checks, endpoint identities, independent replay of saved checkpoints, and numerical comparison against previously saved experimental outputs.
Experimental conclusions were based on results produced by the implemented experiments rather than on generative-AI predictions
Literature and novelty claims arising during AI-assisted discussion were checked against the cited literature.
The authors made the final decisions regarding the mathematical formulation, hypotheses, experimental design, interpretation, claims, and manuscript content.

We take responsibility for the final content of this work, including text, claims, code, analyses, or other artifacts produced with the aid of generative AI.

\section*{Acknowledgments}
This work was partially supported by the Swedish Research Council (VR) through a grant for the project Uncertainty-Aware Transformers for Regression Tasks in Computer Vision (2022-04266). 
Computational resources were provided on the Berzelius system funded by the Knut and Alice Wallenberg foundation and operated by NAISS.

\bibliography{iclr2027_conference}
\bibliographystyle{iclr2027_conference}

\appendix
\section{Proofs}
\label{app:proofs}

\subsection{Equivalence of the predictions and backpropagated gradients}
\label{app:proof_equivalence}

Write
\begin{equation}
    \mathcal H=\frac{\tau}{2}I-\vr\cdot\bm{\sigma},
    \qquad
    \vr=\left(-\nu_2,\nu_3,\frac{\nu_4-\nu_1}{2}\right),
\end{equation}
where $\bm{\sigma}=(\sigma_1,\sigma_2,\sigma_3)$ denotes the Pauli matrices,
\begin{equation}
    \sigma_1=\begin{pmatrix}0&1\\1&0\end{pmatrix},\qquad
    \sigma_2=\begin{pmatrix}0&-i\\i&0\end{pmatrix},\qquad
    \sigma_3=\begin{pmatrix}1&0\\0&-1\end{pmatrix}.
\end{equation}
Since $(\vr\cdot\bm{\sigma})^2=\|\vr\|^2I$, for $\vr\ne0$ the projector onto the
smallest-eigenvalue eigenspace of $\mathcal H$ is
\begin{equation}
    P_-=\frac12\left(I+\frac{\vr}{\|\vr\|}\cdot\bm{\sigma}\right).
\end{equation}
Under the standard identification $\mathbb{CP}^1\simeq\mathcal S^2$, this
projector corresponds to $\vp=\vr/\|\vr\|$. Thus the eigenvector construction and
direct normalization define the same prediction map.

Consequently, for any differentiable task loss $\mathcal L=\ell(\vp)$, both
constructions give
\begin{equation}
    \vg:=\nabla_{\vr}\mathcal L
    =\frac{I-\vp\vp^\top}{\|\vr\|}\nabla_{\vp}\ell,
    \qquad
    \nabla_{\bm{\nu}}\mathcal L
    =\left(-\frac{g_3}{2},-g_1,g_2,\frac{g_3}{2}\right)^\top.
\end{equation}
Their gradients with respect to any shared upstream parameters therefore
agree by the chain rule. Their degeneracies also coincide: $\vr=0$ precisely
when $\lambda_1=\lambda_2$, at which point neither construction determines a
unique sphere prediction.

\paragraph{Parameterization and optimization}
The equivalence is under the stated change of variables. Define
\begin{equation}
    A=\begin{pmatrix}
          0 & -1 & 0 & 0 \\
          0 & 0 & 1 & 0 \\
          -\tfrac12 & 0 & 0 & \tfrac12
    \end{pmatrix},
    \qquad \vr=A\bm{\nu}.
\end{equation}
Then ordinary Euclidean gradient descent directly on $\bm{\nu}$ induces
\begin{equation}
    \Delta\vr=-\eta AA^\top\vg
    =-\eta\,\operatorname{diag}(1,1,\tfrac12)\vg,
\end{equation}
whereas gradient descent directly on $\vr$ gives $-\eta\vg$.

\subsection{The trace receives no task-loss gradient}
\label{app:proof_trace}
The gradient in Appendix~\ref{app:proof_equivalence} satisfies
\begin{equation}
(1,0,0,1) \cdot \nabla_{\bm{\nu}}\mathcal L
=\frac{\partial\mathcal L}{\partial\nu_1}
+\frac{\partial\mathcal L}{\partial\nu_4}=0,
\end{equation}
so it has no component in the trace direction. This does not require the
two diagonal-entry gradients individually to vanish. Nor does it require
the trace output to remain constant when shared upstream features change
during training. This result is intuitive as eigenvalue shifts by some scalar $\gamma$ (i.e. $\mathcal{H} + \gamma I$) do not change the eigenvectors but do change the trace by $2\gamma$.

\section{Experimental details}
\label{app:experimental_details}

All experiments are implemented in PyTorch~\citep{paszke2019pytorch} and run on NVIDIA A100 GPUs.
Each experiment is repeated over five independent training seeds.

\subsection{Surface-normal estimation}
\label{app:surface_normal_details}

\paragraph{Model and task}
We use the baseline-normal architecture from DSINE~\citep{bae2024dsine} on NYUv2~\citep{silberman2012indoor}. The
representation is the 128-dimensional feature vector immediately before the
final $1\times1$ convolution. The final convolution has four output channels.
The first three determine the predicted surface normal after normalization;
the fourth channel is excluded from the prediction and loss and is held fixed
throughout optimization. Training uses the objective in
\cref{eq:dsine_objective}. As in the main analysis, eventual difficulty is the
final model's angular error.

\paragraph{Training and checkpoints}
The model is trained for 20 epochs using AdamW and the optimization settings
of the baseline-normal experiment configuration. Automatic mixed precision is
used on CUDA, and gradients are clipped according to the configured gradient
clipping threshold. The fourth output channel of the final convolution is
restored after every optimizer step so that it remains fixed despite the
optimizer update.
We evaluate the frozen training trajectory at epochs
$\{0,1,2,3,4,5,7,10,15,20\}$. The checkpoint at epoch 0 is the initialized
model before training.

\paragraph{Data and evaluation}
The NYUv2 model-training split contains 795 images. Evaluation uses 327 CAL
images and 327 image-disjoint TEST images. We deterministically sample 83,712
valid pixels from each split. The same sampled pixel locations are used across
training seeds and checkpoints. The eventual-difficulty target is computed
from the final model of the corresponding training seed and held fixed across
the complete checkpoint trajectory.

\subsection{CIFAR-100 classification}
\label{app:classification_details}

\paragraph{Model and task}
We use the \texttt{VisionTransformer} implementation provided by
\texttt{torchvision}, adapted to CIFAR-100 inputs. The model uses
$32\times32$ RGB inputs and $4\times4$ patches, with hidden dimension 192,
12 transformer layers, 3 attention heads, and MLP dimension 768. Dropout and
attention dropout are set to zero. The representation is the 192-dimensional
CLS token immediately before the final $192\rightarrow100$ linear classifier.
Training uses the objective in \cref{eq:classification_objective}. As in the
main analysis, eventual difficulty is the final model's NLL.

\paragraph{Training and checkpoints}
The model is trained for 300 epochs with AdamW, learning rate $10^{-3}$,
weight decay $0.05$, and batch size 256. The learning-rate schedule uses a
10-epoch linear warmup followed by cosine decay. Training augmentation
consists of random $32\times32$ crops with padding 4, random horizontal
flips, and RandAugment with two operations and magnitude 9. Each training
batch then uses either Mixup or CutMix with equal probability, with
$\alpha_{\mathrm{mixup}}=0.2$ and $\alpha_{\mathrm{cutmix}}=1.0$.
Label smoothing is 0.1. CAL and TEST use deterministic preprocessing without
training augmentation.
We evaluate checkpoints at epochs
$\{0,1,2,5,10,20,40,80,120,160,200,250,300\}$, including the initialized
model at epoch 0.

\paragraph{Data and evaluation}
The official CIFAR-100 training set is split deterministically into 45,000
TRAIN examples and 5,000 CAL examples using split seed 424242. The official
10,000-example test set is used as TEST. The same split is used across
training seeds. The eventual-difficulty target is computed from the final
model of the corresponding training seed and held fixed across checkpoints.

\subsection{CIFAR-100 inpainting}
\label{app:inpainting_details}

\paragraph{Model and task}
The inpainting model uses the same \texttt{torchvision}
\texttt{VisionTransformer} implementation and transformer architecture:
$32\times32$ RGB inputs, $4\times4$ patches, hidden dimension 192,
12 transformer layers, 3 attention heads, and MLP dimension 768. Each image
produces 64 spatial tokens. The CLS token passes through the transformer but
is not used by the reconstruction head.

A central $16\times16$ region is masked. Before model normalization, the
masked pixels are replaced by the CIFAR-100 channel mean. A shared
$\operatorname{Linear}(192,48)$ head is applied independently to each spatial
token, where the 48 outputs represent the RGB values of one $4\times4$ patch.
The analyzed representation is the 192-dimensional spatial token immediately
before this reconstruction head. Training uses the objective in \cref{eq:inpainting_objective}. As in the main analysis, eventual difficulty is
the final model's per-patch reconstruction MSE.

\paragraph{Training and checkpoints}
The model is trained for 100 epochs with AdamW, learning rate $10^{-3}$,
weight decay $0.05$, and batch size 256. A 5-epoch linear warmup is followed
by cosine decay. Training augmentation consists of random $32\times32$ crops
with padding 4 and random horizontal flips; the central mask is applied after
the geometric augmentation. CAL and TEST evaluation is deterministic.
We evaluate checkpoints at epochs
$\{0,1,2,5,10,20,40,60,80,100\}$, including the initialized model at epoch 0.

\paragraph{Data and evaluation}
We use the same deterministic 45,000 TRAIN / 5,000 CAL split of the
CIFAR-100 training set as in classification, together with the official
10,000-image TEST set. Each image contributes the 16 spatial tokens
corresponding to the masked region, yielding 80,000 CAL observations and
160,000 TEST observations. The eventual-difficulty target is computed from
the final model of the corresponding training seed and held fixed across
checkpoints.

\subsection{Checkpoint and evaluation details}
\label{app:evaluation_details}
\paragraph{Passive directions}
For every task and training seed, 256 passive directions are generated from
independent zero-mean isotropic Gaussian draws, sampled once and held fixed
across all checkpoints. In the two CIFAR-100 experiments,
$\va_j\sim\mathcal{N}(0,I_d/d)$ with $d=192$. For surface-normal estimation,
the Gaussian probe directions are additionally rescaled to a common norm
matching the weight norm of the fourth output channel in the baseline model's
final $1\times1$ convolution.

The probe directions are independent of the eventual-difficulty property.
No centering is applied as part of the probe construction. Centering is
instead applied where required by the individual statistics, as defined in
the main text.

\paragraph{Fixed task-readout subspace}
For the orthogonal and strict analyses, the task-readout projector defined in
the main text is constructed from the final readout $W$ and held fixed
across all checkpoints. For surface-normal estimation, $W$ contains the
three rows of the final convolution that determine the predicted normal; the
fixed fourth output channel is not part of the task readout. Thus the
task-readout subspace removed by these controls does not change across the
training trajectory.

\paragraph{Calibration and TEST evaluation}
All fitted analysis quantities are estimated using CAL and then frozen before
TEST evaluation. For $A_G$, the bandwidth of each scalar coordinate is its
population standard deviation on CAL at the corresponding checkpoint. The
primary bandwidth is one CAL standard deviation. For calibrated $R^2$, the
primary analysis uses 20 CAL quantile bins. Bin boundaries and the mean
eventual difficulty within each bin are estimated on CAL and then applied
unchanged to TEST.

For surface-normal estimation and inpainting, observations from the same
image are kept together whenever an analysis procedure requires grouping.
TEST is not used to select bandwidths, calibration mappings, learned-probe
regularization, or other fitted analysis quantities.

\section{Passive probe controls}
\label{app:probe_controls}

We evaluate four constructions to separate the effects of feature
normalization and removal of the final task-readout row space. Let
$h_{\bot}(x)=(I-P_W)h(x)$ and $\va_{j,\bot}=(I-P_W)\va_j$, where $P_W$ is the
projector defined in the main text. The four scalar coordinates are
\begin{equation}
    \begin{aligned}
        q^{\mathrm{raw}}_j(x)
        &= \va_j^\top h(x),
        \qquad\qquad
        q^{\mathrm{unit}}_j(x)
        = \va_j^\top\frac{h(x)}{\|h(x)\|}, \\
        q^{\mathrm{orth}}_j(x)
        &= \va_{j,\bot}^\top h_{\bot}(x),
        \qquad
        q^{\mathrm{strict}}_j(x)
        =
        \frac{\va_{j,\bot}^\top h_{\bot}(x)}
        {\|\va_{j,\bot}\|_2\|h_{\bot}(x)\|}.
    \end{aligned}
\end{equation}

Raw applies neither control, unit normalizes the representation only, orth
removes the final task-readout row space, and strict combines readout removal
with normalization.

As mentioned in \cref{app:evaluation_details}, the experiment implementations differ slightly in the fixed radial scaling
assigned to probe directions. Such fixed positive per-probe rescaling does
not affect Pearson squared, absolute Spearman, calibrated $R^2$, or the
one-standard-deviation $A_G$ statistic. The same final task-readout subspace
is used at every checkpoint.

Table~\ref{tab:probe_controls} reports the mean within-seed endpoint change
and its sample standard deviation across the five training seeds. All entries
are multiplied by $10^3$.

\begin{table*}[t]
    \centering
    \caption{Passive-probe control analysis. Entries are mean within-seed endpoint
    change $\Delta$ with sample standard deviation across five training seeds,
        in units of $10^{-3}$.}
    \label{tab:probe_controls}
    \small
    \begin{tabular}{llrrr}
        \toprule
        Task & Probe family & $\Delta A_K$ & $\Delta A_G$ &
        $\Delta R^2_{\mathrm{cal}}$ \\
        \midrule
        Surface normals
        & Raw    & $+7.27\pm5.28$ & $+7.01\pm1.19$ & $+36.37\pm3.98$ \\
        & Unit   & $-0.92\pm1.60$ & $+5.79\pm1.42$ & $+38.12\pm4.92$ \\
        & Orth   & $+9.26\pm4.82$ & $+6.39\pm1.32$ & $+31.90\pm3.77$ \\
        & Strict & $-0.79\pm2.04$ & $+5.45\pm1.33$ & $+34.63\pm4.61$ \\
        \midrule
        Classification
        & Raw    & $-0.11\pm0.37$ & $-0.03\pm0.22$ & $+0.84\pm0.47$ \\
        & Unit   & $-0.19\pm0.36$ & $-0.08\pm0.21$ & $+0.60\pm0.37$ \\
        & Orth   & $+0.05\pm0.55$ & $+0.02\pm0.43$ & $+1.07\pm0.93$ \\
        & Strict & $-0.25\pm0.53$ & $-0.13\pm0.37$ & $+0.32\pm0.62$ \\
        \midrule
        Inpainting
        & Raw    & $+3.88\pm0.87$ & $+2.36\pm0.55$ & $+4.54\pm1.11$ \\
        & Unit   & $+3.93\pm0.88$ & $+2.35\pm0.56$ & $+4.55\pm1.12$ \\
        & Orth   & $+8.00\pm1.52$ & $+3.98\pm0.70$ & $+8.35\pm1.43$ \\
        & Strict & $+8.73\pm1.61$ & $+4.30\pm0.79$ & $+9.42\pm1.61$ \\
        \bottomrule
    \end{tabular}
\end{table*}

The qualitative task-level patterns are unchanged across the control
constructions. Surface-normal estimation shows positive nonlinear and
calibrated changes when normalization or readout removal is applied separately
or jointly, classification remains close to its initialization level, and
inpainting shows positive changes throughout. For surface normals, the
near-zero strict $A_K$ change arises when readout removal and normalization
are combined; the corresponding $A_G$ and calibrated-$R^2$ changes remain
positive.

\section{Calibration sensitivity}
\label{app:bin_ablation}

The primary calibrated-$R^2$ analysis uses 20 CAL quantile bins. To assess
sensitivity to this choice, we repeat the endpoint analysis with
$B\in\{5,10,20,40\}$ bins while leaving the probe coordinates and
eventual-difficulty targets unchanged. Table~\ref{tab:bin_ablation} reports
raw and strict probes, the two families used in the main analysis.

\begin{table*}[t]
    \centering
    \caption{Sensitivity of calibrated $R^2$ to the number of CAL quantile bins.
    Entries are mean within-seed endpoint change $\Delta R^2_{\mathrm{cal}}$ with
    sample standard deviation across five training seeds, in units of $10^{-3}$.}
    \label{tab:bin_ablation}
    \small
    \begin{tabular}{llrrrr}
        \toprule
        Task & Probe family & $B=5$ & $B=10$ & $B=20$ & $B=40$ \\
        \midrule
        Surface normals
        & Raw
        & $+30.52\pm3.92$
        & $+34.60\pm3.85$
        & $+36.37\pm3.98$
        & $+37.11\pm4.01$ \\
        & Strict
        & $+23.27\pm3.34$
        & $+30.59\pm4.19$
        & $+34.63\pm4.61$
        & $+36.39\pm4.79$ \\
        \midrule
        Classification
        & Raw
        & $+0.34\pm0.40$
        & $+0.57\pm0.46$
        & $+0.84\pm0.47$
        & $+0.98\pm0.45$ \\
        & Strict
        & $+0.09\pm0.57$
        & $+0.23\pm0.60$
        & $+0.32\pm0.62$
        & $+0.29\pm0.63$ \\
        \midrule
        Inpainting
        & Raw
        & $+4.25\pm1.04$
        & $+4.41\pm1.09$
        & $+4.54\pm1.11$
        & $+4.62\pm1.12$ \\
        & Strict
        & $+8.29\pm1.48$
        & $+9.02\pm1.57$
        & $+9.42\pm1.61$
        & $+9.60\pm1.63$ \\
        \bottomrule
    \end{tabular}
\end{table*}

The endpoint conclusions are stable across the tested bin counts.
Surface-normal estimation and inpainting retain positive calibrated-$R^2$
changes under both raw and strict probes, whereas classification remains
close to zero. The magnitude increases somewhat with finer calibration for
surface normals, but the cross-task pattern does not depend on the primary
choice $B=20$.

\section{Learned-probe protocol}
\label{app:learned_probe_details}

For each task, training seed, and checkpoint, we fit a separate ridge
regression from the raw representation to the same eventual-difficulty target
used in the main analysis. This provides the learned-probe quantities reported
in the main-text comparison.

Let $\vz_i$ denote the raw representation after feature-wise standardization
estimated from the relevant CAL training fold. The learned scalar coordinate
is
\begin{equation}
    \hat e_i=\vb^\top\vz_i+c~,
\end{equation}
where $\vb$ and $c$ are fitted by minimizing squared error with an
$\ell_2$ penalty on $\vb$. The intercept is not regularized.

\paragraph{Regularization selection}
The ridge parameter is selected independently for each training seed and
checkpoint by five-fold cross-validation on CAL using
$\lambda\in\{10^{-6},10^{-5},10^{-4},10^{-3},10^{-2},10^{-1},
1,10,100,1000\}$. Feature means and standard deviations are estimated within
each cross-validation training fold and applied to its held-out fold, so
held-out CAL observations do not contribute to their own standardization.

For surface-normal estimation, folds are grouped by image so that pixels from
the same image cannot appear in both the fitting and validation portions of a
fold. Inpainting is treated analogously: all 16 masked-patch observations from
one image are assigned to the same fold. Classification uses ordinary
five-fold splitting over images.

After selecting $\lambda$, feature standardization and the ridge model are
refit using the complete CAL set and evaluated once on TEST. The
regularization parameter is allowed to differ across checkpoints because the
learned observer is permitted to adapt independently to each representation.

\paragraph{Matched evaluation}
Pearson squared and absolute Spearman are computed directly between the
learned scalar prediction $\hat e$ and eventual difficulty on TEST. For the
matched calibrated-$R^2$ comparison, the learned scalar is passed through the
same 20-bin CAL-to-TEST calibration procedure used for passive probes:
quantile-bin boundaries and within-bin target means are estimated from CAL,
frozen, and then evaluated on TEST.

TEST observations are not used for feature standardization, ridge fitting,
regularization selection, or scalar calibration. Direct ridge TEST $R^2$ is
treated as a diagnostic quantity and is not included in the main
passive-versus-learned comparison.

\end{document}